\pdfoutput=1
\documentclass[11pt]{article}

\usepackage[final]{acl}

\usepackage{times}
\usepackage{latexsym}
\usepackage{booktabs}
\usepackage{multirow}
\usepackage{pifont}
\usepackage{float}
\usepackage{graphicx}
\usepackage{subfigure}
\usepackage{caption}
\usepackage{mathtools}
\usepackage{booktabs}
\usepackage{multirow}
\usepackage{ragged2e}
\usepackage{hyperref}
\usepackage{multicol}
\usepackage{siunitx}
\usepackage{adjustbox}
\usepackage{pifont}
\usepackage{tabularx}
\usepackage{enumitem}
\usepackage{pythonhighlight}
\usepackage{array, makecell, multirow}
\newcolumntype{P}[1]{>{\centering\arraybackslash}p{#1}}
\usepackage{siunitx}
\usepackage{soul}
\usepackage{xcolor}
\usepackage{colortbl}
\sethlcolor{cYellow}
\usepackage{tcolorbox}
\usepackage{amsmath}
\usepackage{amssymb}% http://ctan.org/pkg/amssymb
\usepackage{pifont}

\newcolumntype{R}[1]{>{\RaggedLeft\arraybackslash}p{#1}}
\newcolumntype{L}[1]{>{\RaggedRight\arraybackslash}p{#1}}
\tcbuselibrary{skins, breakable, theorems}
\usepackage[T1]{fontenc}
\usepackage[utf8]{inputenc}

\usepackage{microtype}

\usepackage{inconsolata}
\usepackage{float}
\usepackage{graphicx}
\usepackage{fontawesome5}
\title{From Automation to Autonomy:\\ A Survey on Large Language Models in Scientific Discovery}

\author{Tianshi Zheng, Zheye Deng, Hong Ting Tsang, Weiqi Wang,\\ \textbf{Jiaxin Bai, Zihao Wang, Yangqiu Song}\\
Department of Computer Science and Engineering, HKUST, Hong Kong SAR, China\\
\href{https://github.com/HKUST-KnowComp/Awesome-LLM-Scientific-Discovery}{\texttt{https://github.com/HKUST-KnowComp/Awesome-LLM-Scientific-Discovery}}}

\begin{document}
\maketitle
\begin{abstract}
Large Language Models (LLMs) are catalyzing a paradigm shift in scientific discovery, evolving from task-specific automation tools into increasingly autonomous agents and fundamentally redefining research processes and human-AI collaboration. This survey systematically charts this burgeoning field, placing a central focus on the changing roles and escalating capabilities of LLMs in science. Through the lens of the scientific method, we introduce a foundational three-level taxonomy—\textit{Tool, Analyst, and Scientist}—to delineate their escalating autonomy and evolving responsibilities within the research lifecycle. We further identify pivotal challenges and future research trajectories such as robotic automation, self-improvement, and ethical governance. Overall, this survey provides a conceptual architecture and strategic foresight to navigate and shape the future of AI-driven scientific discovery, fostering both rapid innovation and responsible advancement.
\end{abstract}

\section{Introduction}

The relentless advancement of Large Language Models (LLMs) has unlocked a suite of emergent abilities, such as planning \cite{huang2024understandingplanningllmagents}, complex reasoning \cite{huang-chang-2023-towards}, and instruction following \cite{qin2024infobenchevaluatinginstructionfollowing}. Moreover, integrating agentic workflows enables LLM-based systems to perform advanced functions, including web navigation \cite{he2024webvoyagerbuildingendtoendweb}, tool use \cite{Qu_2025}, code execution \cite{jiang2024surveylargelanguagemodels}, and data analytics \cite{sun2024surveylargelanguagemodelbased}. In scientific discovery, this convergence of advanced LLM capabilities and agentic functionalities is catalyzing a significant paradigm shift. This shift is poised not only to accelerate the research lifecycle but also to fundamentally alter the collaborative dynamics between human researchers and artificial intelligence in the pursuit of knowledge.

\begin{figure}[]
\begin{center}
\includegraphics[clip,width=230pt]{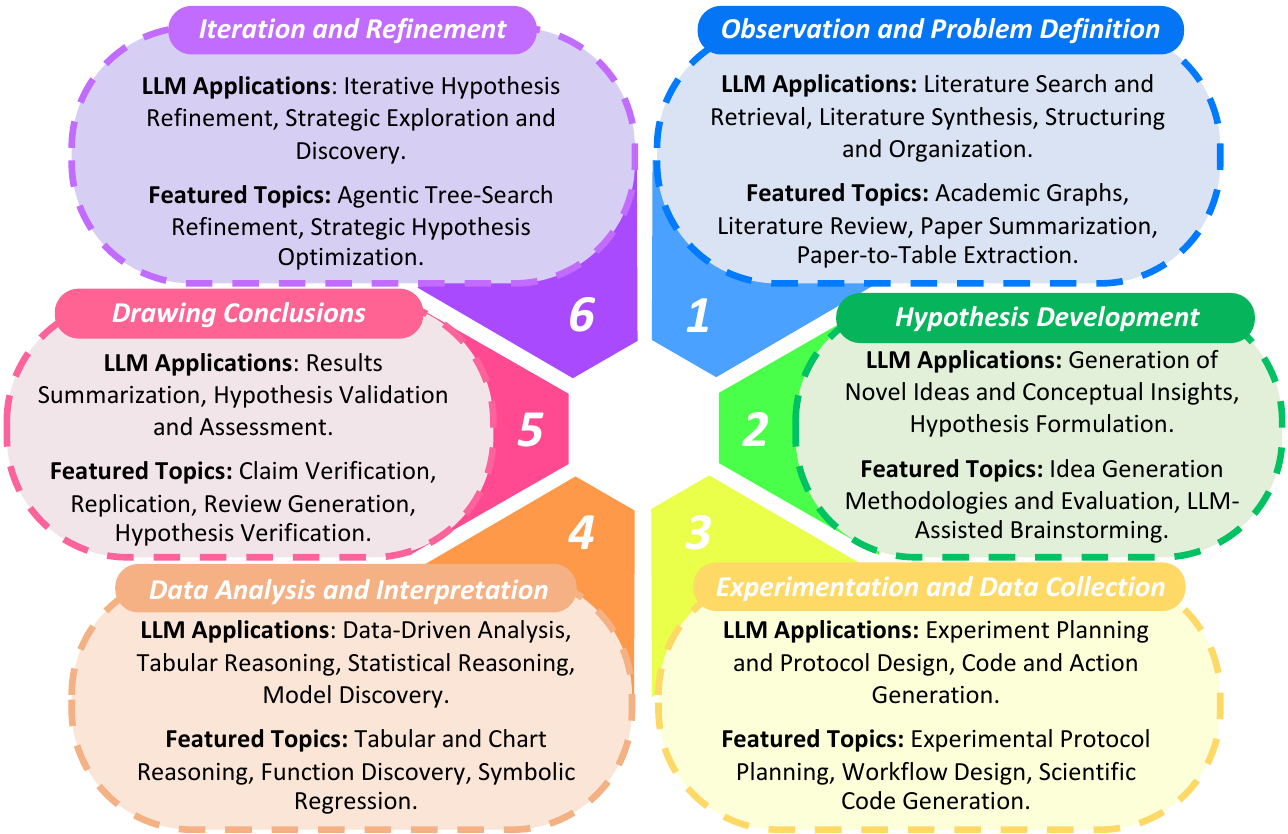}
\end{center}
\caption{Stages of the scientific method with corresponding LLM applications and research topics.}
\vspace{-0.2cm}
\label{fig:scimethod}
\end{figure}

\begin{table*}[t]
\centering
\small
\begin{tabular}{ccccc}
\toprule
\textbf{Autonomy Levels} & \textbf{LLMs' Role} & \textbf{Human's Role} & \textbf{Task Scope} & \textbf{Agentic Workflow} \\ \midrule
\textbf{\begin{tabular}[c]{@{}c@{}}\faTools\ Level 1\\ \textit{LLM as Tool}\end{tabular}} & Task Automation Tool & Task Allocation & Explicitly Defined & Simple \& Static \\ \midrule
\textbf{\begin{tabular}[c]{@{}c@{}}\faChartBar\ Level 2\\ \textit{LLM as Analyst}\end{tabular}} & \begin{tabular}[c]{@{}c@{}} Data Modeling \&\\ Analytical Agent \end{tabular} & \begin{tabular}[c]{@{}c@{}}Problem Definition \&\\ Output Validation\end{tabular}  & Goal-Oriented & Advanced \\ \midrule
\textbf{\begin{tabular}[c]{@{}c@{}}\faLightbulb\ Level 3\\ \textit{LLM as Scientist}\end{tabular}} & \begin{tabular}[c]{@{}c@{}}Open Exploratory \&\\ Discovery Agent \end{tabular} & Minimal Intervention & Open-Ended & Strategic \& Iterative \\ \bottomrule
\end{tabular}
\caption{Three levels of autonomy in LLM-based scientific discovery.}
\vspace{-0.2cm}
\label{tab:levels}
\end{table*}

However, this rapid expansion of LLM applications and the ongoing paradigm shift in scientific discovery present notable challenges. The accelerated pace of LLM evolution and their deepening integration into complex research complicate systematic assessment, necessitating conceptual frameworks to structure current understanding and chart future directions. While existing surveys have provided valuable overviews of LLMs in various scientific domains \cite{zhang2024comprehensivesurveyscientificlarge, 10.1145/3715318} or have cataloged particular AI techniques for science \cite{luo2025llm4srsurveylargelanguage, Reddy_Shojaee_2025}, they often focus on discipline-specific applications or a static snapshot of LLM capabilities. Consequently, existing reviews may overlook the crucial trend of increasing LLM autonomy and their evolving roles across the entire scientific method, leaving their comprehensive impact and trajectory towards greater independence underexplored.

To systematically chart this evolving landscape and address the identified gap, we anchor our analysis in the six stages (Figure \ref{fig:scimethod}) of the established scientific method \cite{Popper1935-POPTLO-7, Kuhn1962-KUHTSO-3}: (1) observation and problem definition, (2) hypothesis development, (3) experimentation and data collection, (4) data analysis and interpretation, (5) drawing conclusions, and (6) iteration and refinement. Our examination of LLM applications across these stages reveals a significant trend: LLMs are progressing from performing discrete, task-oriented functions within a single stage to deployment in sophisticated, multi-stage agentic workflows. Notably, emerging research \cite{schmidgall2025agentlaboratoryusingllm,yamada2025aiscientistv2workshoplevelautomated} now explores developing LLM-based systems capable of autonomously navigating nearly all these stages. To effectively capture and delineate this trajectory of increasing capability and independence, we introduce a foundational three-level taxonomy for LLM involvement in scientific discovery (Table \ref{tab:levels}): (i) \textbf{\textit{LLM as Tool}}, where models augment human researchers by performing specific, well-defined tasks under direct supervision; (ii) \textbf{\textit{LLM as Analyst}}, where models exhibit greater autonomy in processing complex information, conducting analyses, and offering insights with reduced human intervention; and (iii) \textbf{\textit{LLM as Scientist}}, representing a more advanced stage where LLM-based systems can autonomously conduct major research stages, from formulating hypotheses to interpreting results and suggesting new avenues of inquiry.

Building upon this taxonomic framework, we further identify critical gaps in the current research landscape and highlight pivotal challenges and future trajectories for the field, including: (1) fully autonomous discovery cycles for evolving scientific inquiry without human intervention; (2) robotic automation for interaction in the physical world for laboratory experimentation; (3) continuous self-improvement and adaptation from past research experiences; (4) transparency and interpretability of LLM-conducted research; and (5) ethical governance and societal alignment. Addressing these multifaceted challenges will be crucial for achieving a future where AI acts as a transformative partner in scientific exploration.

This survey focuses on LLM-based systems in scientific discovery, particularly their varying levels of autonomy. While acknowledging the broad impact of LLMs in science, we deliberately narrow our scope to exclude research on general-purpose scientific LLMs or LLMs for domain-specific scientific knowledge acquisition and reasoning, which are well covered in existing surveys \cite{zhang2024comprehensivesurveyscientificlarge, 10.1145/3715318}. The remainder of this paper is organized as follows: Section \ref{sec:levels} details our taxonomy and its interaction with the scientific method. Section \ref{sec:l1} presents \textit{LLM as Tool} applications, categorized by scientific method stages. Section \ref{sec:l2} examines \textit{LLM as Analyst} works by scientific domain, while Section \ref{sec:l3} analyzes \textit{LLM as Scientist} systems, focusing on their idea development and refinement strategies. Section \ref{sec:future} explores challenges and future directions.

\begin{figure*}[t]
\begin{center}
\includegraphics[clip,width=425pt]{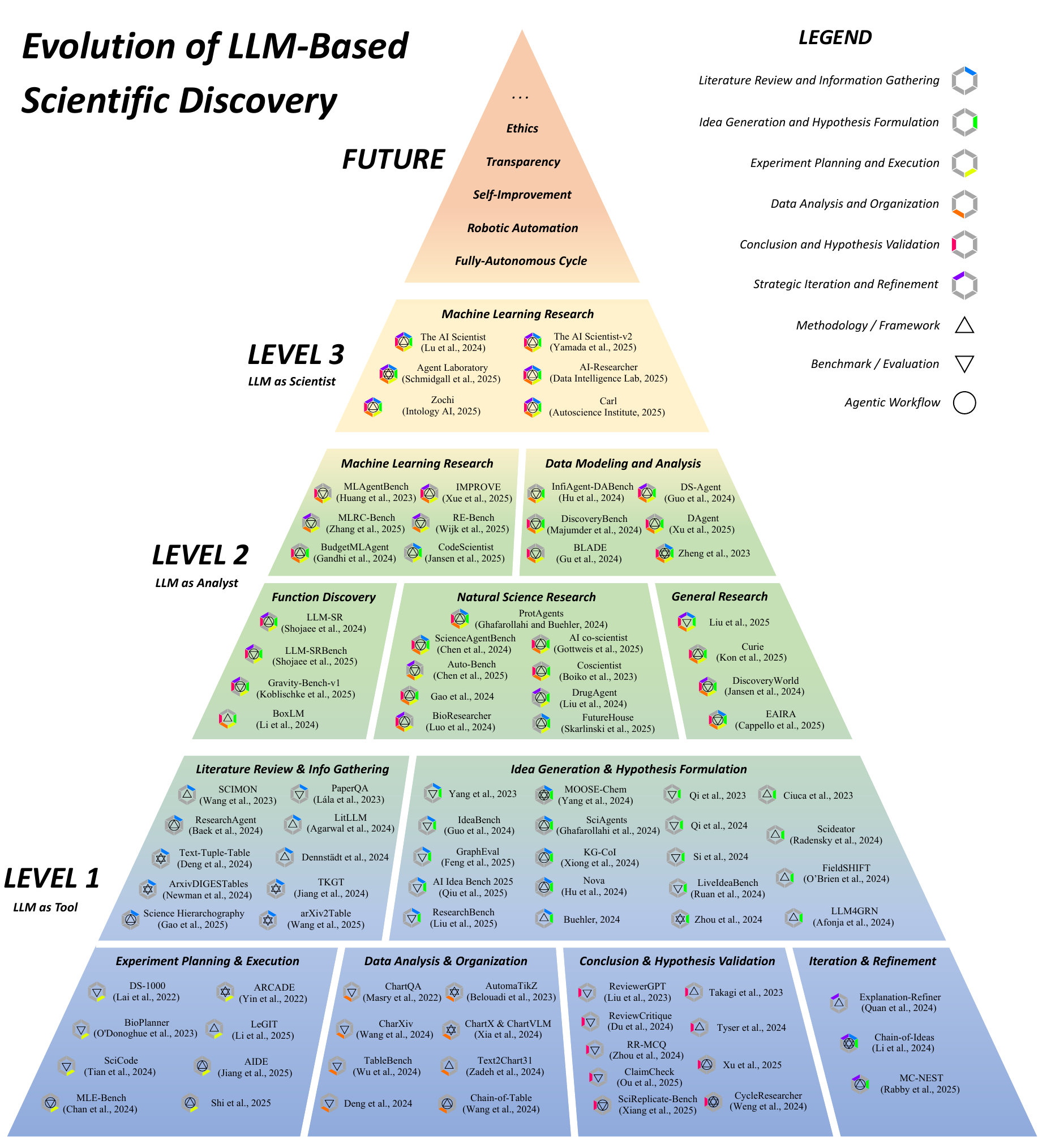}
\end{center}
\caption{Taxonomy of research works in LLM-based scientific discovery with detailed categorization.}
\label{fig:main_taxonomy}
\vspace{-0.2cm}
\end{figure*}
\section{Three Levels of Autonomy}
\label{sec:levels}
Table \ref{tab:levels} illustrates the three levels of autonomy in LLM-based scientific discovery with their associated features. In this section, we discuss their applications and characteristics in more detail.

\vspace{0.1cm}
\noindent\textbf{LLM as Tool (Level 1).}
Level 1 represents the most foundational application of LLMs in scientific discovery. At this stage, LLMs function primarily as \textit{tailored tools} under direct human supervision, designed to execute specific, well-defined tasks within a single stage of the scientific method. Their role is to augment human capabilities by automating or accelerating discrete activities such as literature summarization, drafting initial text for manuscripts, generating code snippets for data processing, or reformatting datasets. The autonomy of LLMs at this level is limited; they operate based on explicit human prompts and instructions, with outputs typically requiring human validation and integration into the broader research workflow. The primary goal is to enhance researcher efficiency and reduce routine task burdens.

\vspace{0.1cm}
\noindent\textbf{LLM as Analyst (Level 2).}
In Level 2, LLMs exhibit a greater degree of autonomy and move beyond purely static, task-oriented applications. Here, LLMs function as \textit{passive agents}, capable of more complex information processing, data modeling, and analytical reasoning with reduced human intervention for intermediate steps. While still operating within boundaries set by human researchers, these systems can independently manage sequences of tasks, such as analyzing experimental datasets to identify trends, interpreting outputs from complex simulations, or even performing iterative refinement of models. The human researcher typically defines the overall analytical goals, provides the necessary data, and critically evaluates the insights or interpretations generated by the LLM.

\vspace{0.1cm}
\noindent\textbf{LLM as Scientist (Level 3).}
Level 3 applications signify a significant leap in autonomy, where LLM-based systems operate as \textit{active agents} capable of orchestrating and navigating multiple stages of the scientific discovery process with considerable independence. These systems can demonstrate initiative in formulating hypotheses, planning and executing experiments, analyzing the resultant data, drawing preliminary conclusions, and potentially proposing subsequent research questions or avenues for exploration. LLM-based systems at this level can drive substantial portions of the research cycle, conducting scientific discovery with minimal human intervention.

\vspace{0.1cm}
Collectively, we present our full taxonomy with detailed categorization in Figure \ref{fig:main_taxonomy}, which consolidates research works within our focused scope across all three levels of autonomy.

\section{Level 1. LLM as Tool (\autoref{tab:level1})}  
\label{sec:l1}
In this section, we introduce \textit{Level 1} research works in LLM-based scientific discovery, categorized by the stages in the scientific method they address.

\subsection{Literature Review and Information Gathering}
\label{sec:l1-1}

\paragraph{Literature Review} Automatic literature search and retrieval is crucial for identifying research gaps and formulating research questions. \citet{Lala2023PaperQA} first introduced the literature retrieval benchmark LitQA, featuring a RAG-based agent, PaperQA. LitLLM \cite{agarwal2024litllmtoolkitscientificliterature} further provided a comprehensive RAG-based toolkit for LLM-driven literature review. Taking this automation a step further, \citet{wang2024autosurvey} demonstrated that large language models can automatically write entire survey papers. \citet{Dennstadt2024Title} directed their focus to the biomedical domain, highlighting the potential of LLMs in literature screening. Other approaches, such as SCIMON \cite{wang2024scimonscientificinspirationmachines} and ResearchAgent \cite{baek2025researchagentiterativeresearchidea}, have integrated active literature retrieval with the generation of research ideas. More recently, \citet{gao2025sciencehierarchographyhierarchicalorganization} tackled the task of hierarchically organizing scientific literature through fine-grained paper abstraction. Nevertheless, several `Deep Research' products \cite{OpenAI2025deepresearch,GoogleGemini2025deepresearch,xAI2025grok3} have recently been released, featuring enhanced agentic workflows that support automated literature web search, organization, and report generation, thereby significantly accelerating traditional, human-intensive literature research processes.

\paragraph{Information Aggregation} In parallel, several works have explored effective methods for aggregating information from scientific papers into tabular summaries. ArxivDIGESTables \cite{newman2024arxivdigestablessynthesizingscientificliterature} investigated cross-literature table generation using LLMs, accompanied by an automated evaluation strategy. ArXiv2Table \cite{wang2025llmsgeneratetabularsummaries} revised the evaluation protocol and provided a comprehensive benchmark. Methodologies such as Text-Tuple-Table \cite{deng2024texttupletableinformationintegrationtexttotable} and TKGT \cite{jiang-etal-2024-tkgt} have enhanced the quality of LLM-based table generation by incorporating tuple-based structures and graph modalities.

\subsection{Idea Generation and Hypothesis Formulation}
\label{sec:l1-2}
\paragraph{Idea Generation} Numerous research efforts have focused on the automated generation of novel research ideas and hypotheses. In the general domain, benchmarks such as IdeaBench \cite{guo2024ideabenchbenchmarkinglargelanguage} and LiveIdeaBench \cite{ruan2025liveideabenchevaluatingllmsdivergent} have evaluated the capability of LLMs to generate research ideas based on provided literature summaries. Concurrently, LLM-based agent frameworks, including Nova \cite{hu2024novaiterativeplanningsearch}, SciAgents \cite{ghafarollahi2024sciagentsautomatingscientificdiscovery}, and KG-CoI \cite{xiong2024improvingscientifichypothesisgeneration}, have been proposed to enhance idea generation through effective reasoning over academic knowledge graphs, iterative planning, and searching. More specific methodologies, such as employing dynamic control to guide the creative process, have also been introduced \cite{li2024learning}. Moreover, several exploratory studies have assessed the novelty and quality of LLM-generated ideas for AI research, underscoring the potential for automated idea generation when coupled with appropriate human guidance \cite{si2024llmsgeneratenovelresearch, feng2025graphevallightweightgraphbasedllm, qiu2025aiideabench2025}.
Furthermore, many studies within natural science disciplines have investigated LLM-based idea generation in domain-specific contexts. For example, \citet{ciucă2023harnessingpoweradversarialprompting} proposed adopting adversarial prompting for effective idea generation in astronomy. In biology, \citet{buehler2024acceleratingscientificdiscoverygenerative} enhanced idea generation by integrating knowledge extraction and graph representations.

\paragraph{Hypothesis Formulation} Building upon identified ideas, the design of testable scientific hypotheses has also been a significant focus. \citet{qi2023largelanguagemodelszero} and \citet{yang2024largelanguagemodelsautomated} examined the ability of LLMs to propose hypotheses, demonstrating their considerable capacity for generating novel yet valid hypotheses under open-ended constraints. Methodologies such as Scideator \cite{radensky2025scideatorhumanllmscientificidea} have been developed to investigate human-LLM collaboration to facilitate grounded idea and hypothesis generation. Other approaches have focused on ensuring the generated hypotheses are well-founded; for instance, HypER generates literature-grounded hypotheses with clear provenance \cite{vasu2025hyper}, while \citet{oneill2025sparks} leverage structured data from scientific papers for the same purpose. Within natural science, benchmarks \cite{qi2024largelanguagemodelsbiomedical} and methods \cite{D3DD00185G} have extended hypothesis generation into the biomedical domain. Meanwhile, MOOSE-Chem \cite{yang2025moosechemlargelanguagemodels} offers a systematic evaluation benchmark and an agent framework specifically for hypothesis discovery in chemistry.

\subsection{Experiment Planning and Execution}
\label{sec:l1-3}
Experiment planning and execution constitute a crucial stage in LLM-based scientific discovery. While integral to advanced \textit{Level 2} and \textit{Level 3} agents, this subsection focuses on \textit{Level 1} research, where LLMs serve as tools for experimental tasks.

\paragraph{Planning} Regarding experiment planning, \citet{li2025largelanguagemodelshelp} discussed the effectiveness of incorporating LLMs into the design of causal discovery experiments. BioPlanner \cite{odonoghue2023bioplannerautomaticevaluationllms} introduced an automated evaluation framework for assessing LLMs in biological protocol planning. Furthermore, \citet{shi2025hierarchicallyencapsulatedrepresentationprotocol} proposed a hierarchically encapsulated representation to complement LLMs in biological protocol design.

\paragraph{Execution} For experiment execution, current research has primarily concentrated on code generation, particularly for artificial intelligence research, given the inherent compatibility of terminal interfaces with LLM experimental environments. Early code generation benchmarks, such as ARCADE \cite{yin2022naturallanguagecodegeneration} and DS-1000 \cite{lai2022ds1000naturalreliablebenchmark}, focused on data science tasks. Subsequent works, including MLE-Bench \cite{chan2025mlebenchevaluatingmachinelearning} and SciCode \cite{tian2024scicoderesearchcodingbenchmark}, incorporate more realistic scenarios, such as those encountered in machine learning engineering and natural science research, thereby presenting significant challenges for LLMs. To address these challenges, AIDE \citep{jiang2025aideaidrivenexplorationspace} proposed enhancing complex code generation capabilities by adopting tree-search methodologies for code optimization.

\subsection{Data Analysis and Organization}
\label{sec:l1-4}

\paragraph{Tabular Data} In this stage, LLMs assist the scientific workflow by automating processes related to data organization, presentation, and analysis. For data presented in tabular format, Chain-of-Table \cite{wang2024chainoftableevolvingtablesreasoning} proposes a method to enhance tabular understanding by incorporating evolving tables within the reasoning chain of LLMs. Concurrently, \citet{deng2024tablestextsimagesevaluating} highlight the potential of integrating visual information to improve multimodal understanding, thereby aiding tabular comprehension. More recently, \citet{wu2025tablebenchcomprehensivecomplexbenchmark} introduced TableBench, a comprehensive benchmark for table-based question answering under practical industrial scenarios.

\paragraph{Chart Data} Beyond tabular data, charts represent another important format for organizing and storing information derived from experimental data. Early benchmarks, exemplified by ChartQA \cite{masry2022chartqabenchmarkquestionanswering}, examined the capabilities of vision transformers in chart-based question answering. Subsequent works, including CharXiv \cite{wang2024charxivchartinggapsrealistic} and ChartX \cite{xia2025chartxchartvlmversatile}, have expanded the scope of chart understanding scenarios by utilizing human-curated chart generation or by incorporating real-world chart data sourced from arXiv preprints. Regarding chart generation, AutomaTikZ \cite{belouadi2024automatikztextguidedsynthesisscientific} formulates the process as TikZ code generation from caption text and has demonstrated the efficacy of fine-tuning LLMs using open scientific figure data. More recently, Text2Chart31 \cite{zadeh2025text2chart31instructiontuningchart} employed reinforcement learning with automated feedback to refine chart generation capabilities within the Matplotlib library.

\subsection{Conclusion and Hypothesis Validation}
\label{sec:l1-5}

In the concluding stages of research, LLMs can provide feedback on, or verify, claims and conclusions derived from experiments.

\paragraph{Paper Review} In this context, a significant focus of contemporary research involves investigating the utility of LLMs as reviewers for artificial intelligence papers. ReviewerGPT \cite{liu2023reviewergptexploratorystudyusing} initially explored the capability of LLMs to identify deliberately inserted errors within research papers, highlighting the necessity for more robust systems to conduct comprehensive reviews. \citet{zhou-etal-2024-llm} further evaluated static LLMs in the context of reviewing real-world conference papers using a multiple-choice format. \citet{du2024llmsassistnlpresearchers} conducted a comprehensive analysis of LLM review quality through extensive human studies and comparisons, revealing weaknesses in their ability to identify deficiencies. ClaimCheck \cite{ou2025claimcheckgroundedllmcritiques} further investigated the capabilities of LLMs in critiquing research claims, demonstrating that this task remains challenging even for highly advanced models such as OpenAI's o1 \cite{openai2024o1preview}. Beyond reviewing, other work has focused on the subsequent step of paper revision, with systems like XtraGPT enabling human-AI collaboration for controllable academic paper revisions \cite{chen2025xtragpt}. Concurrently, research highlights the potential to address these limitations by incorporating multi-agent systems with specialized roles \cite{tyser2024aidrivenreviewsystemsevaluating, xu2025multiagentreasoningsystemscollaborative}, through LLM alignment via reinforcement learning \cite{weng2025cycleresearcherimprovingautomatedresearch}, or by employing novel frameworks like generative adversarial reviews \cite{bougie2024generative}.

\paragraph{Hypothesis Validation} Another important application at this stage is the automatic validation of hypotheses by LLMs. \citet{takagi2023autonomoushypothesisverificationlanguage} demonstrated that LLMs possess considerable capabilities in automatically generating code to verify research hypotheses within simplified machine learning problems. Benchmarks such as SciReplicate-Bench \cite{xiang2025scireplicatebenchbenchmarkingllmsagentdriven} and PaperBench \cite{starace2025paperbench} have further extended this concept to evaluating the replication of real-world research papers. A distinct but related line of inquiry explores predicting empirical AI research outcomes directly with language models, assessing whether LLMs can anticipate experimental results without full execution \cite{wen2025predicting}. Furthermore, \citet{xu2025advancingaiscientistunderstandingmaking} have navigated this domain into physics research, aiming to enhance the interpretability of the discovery process through the use of multi-agent workflows.
\subsection{Iteration and Refinement}
\label{sec:l1-6}

The iterative refinement of research hypotheses, as a distinct area of investigation, has received comparatively less attention in current research. Explanation-Refiner \cite{quan2024verificationrefinementnaturallanguage} employed theorem provers to verify and subsequently refine LLM-generated hypotheses. Chain-of-Idea \cite{li2024chainideasrevolutionizingresearch} introduced an LLM-based agent framework designed to organize literature and develop research ideas by building upon or extending existing lines of inquiry. More recently, MC-NEST \cite{rabby2025iterativehypothesisgenerationscientific} adopted Monte-Carlo Tree Search to iteratively verify and refine scientific hypotheses across multiple research domains.

\section{Level 2: LLM as Analyst (\autoref{tab:level2})}  
\label{sec:l2}

In this section, we introduce \textit{Level 2} research works in LLM-based scientific discovery, categorized according to their task nature and domains.

\subsection{Machine Learning Research}
\label{sec:l2-1}
Automated Machine Learning (AutoML) \cite{shen2024automatedmachinelearningprinciples} endeavors to generate high-performing modeling configurations for a given task in a data-driven manner. With the advent of LLM-based agents, several studies have explored their application in the automated modeling of machine learning (ML) tasks. A suite of benchmarks has emerged to track progress in this area. MLAgentBench \cite{huang2024mlagentbenchevaluatinglanguageagents} evaluates the capabilities of LLMs in designing and executing ML experiments, revealing that performance is often contingent upon task familiarity. Similarly, MLRC-Bench \cite{zhang2025mlrcbench} and RE-Bench \cite{wijk2024rebench} further probe the limits of these agents, assessing their ability to solve novel ML research challenges and comparing their R\&D capabilities against human experts. MLGym \cite{nathani2025mlgym} offers valuable resource and benchmark for advancing these AI research agents.

To address the challenges posed by these benchmarks, various agentic frameworks have been proposed. The IMPROVE framework \cite{xue2025improveiterativemodelpipeline} highlighted the significance of iterative refinement mechanisms. CodeScientist \cite{jansen2025codescientistendtoendsemiautomatedscientific} incorporated an ML modeling agent with machine-generated ideas, while BudgetMLAgent \cite{gandhi2025budgetmlagentcosteffectivellmmultiagent} leveraged curated expert collaboration frameworks to achieve superior results with cost-effective models. More recent end-to-end systems like MLR-Copilot \cite{li2024mlrcopilot} and the multi-agent framework MLZero \cite{fang2025mlzero} aim for fully autonomous machine learning research and automation. Pushing the boundaries of automation even further, some work has explored the use of language models to directly propose LM architectures \cite{cheng2025languagemodelinglanguagemodels}, moving beyond orchestration to direct model creation.

\subsection{Data Modeling and Analysis}
\label{sec:l2-2}

Automated data-driven analysis, encompassing statistical data modeling and hypothesis validation, represents a foundational application area for LLM-assisted scientific discovery. InfiAgent-DABench \cite{hu2024infiagentdabenchevaluatingagentsdata} benchmarked the capabilities of LLMs in static code generation and execution for data analysis using CSV files. Subsequent benchmarks, such as BLADE \cite{gu2024bladebenchmarkinglanguagemodel}, DiscoveryBench \cite{majumder2024discoverybenchdatadrivendiscoverylarge}, and DSBench \cite{jing2024dsbench}, have improved evaluation robustness by incorporating real-world research papers and expert-curated analytics to assess how far agents are from human expert performance. These studies indicate that most LLMs struggle with complex data analytics tasks, even when operating within an agent framework \cite{zheng2023largelanguagemodelsscientific}. To address these challenges, DS-Agent \cite{guo2024dsagentautomateddatascience} proposes to enhance LLM performance by incorporating a case-based reasoning method to improve domain knowledge acquisition. In a related effort, DAgent \cite{xu2025dagentrelationaldatabasedrivendata} extended the application domain to querying relational databases and enabled report generation using results derived from decomposed sub-problems.
\subsection{Function Discovery} % Assuming this title or similar
\label{sec:l2-3}

Function discovery, which aims to identify the underlying equations from observational data of variables, has been significantly influenced by the advancement of AI-driven symbolic regression (SR) \cite{udrescu2020aifeynmanphysicsinspiredmethod,kamienny2022endtoendsymbolicregressiontransformers}. To enhance this process, LLM-SR \cite{shojaee2025llmsrscientificequationdiscovery} leveraged the prior domain knowledge of LLMs and incorporated feedback from clustered memory storage, while DrSR \cite{wang2025drsr} proposed a dual reasoning framework that utilizes both data and experience for scientific equation discovery. To systematically assess these capabilities, LLM-SRBench \cite{shojaee2025llmsrbenchnewbenchmarkscientific} introduced a benchmark for evaluating LLMs as function discovery agents, which incorporates function transformations to mitigate data contamination. Furthermore, other studies have explored the capabilities of LLMs in discovering complex models within specific domains, such as Physics \cite{koblischke2025gravitybenchv1benchmarkgravitationalphysics}, Statistics \cite{li2024automatedstatisticalmodeldiscovery}, and automated neural scaling law discovery \cite{wei2025evosld}.

\subsection{Natural Science Research}
\label{sec:l2-4}

Research has largely focused on applying LLMs to autonomous research workflows for natural science discovery. Auto-Bench \cite{chen2025autobenchautomatedbenchmarkscientific} evaluated LLMs on chemistry and social science tasks based on causal graph discovery, revealing that LLMs perform effectively only when task complexity is highly limited. In contrast, ScienceAgentBench \cite{chen2025scienceagentbenchrigorousassessmentlanguage} provided a multidisciplinary benchmark for LLMs operating within agent frameworks such as CodeAct \cite{wang2024executablecodeactionselicit} and self-debug \cite{chen2023teachinglargelanguagemodels}. This benchmark highlighted the necessity for tailored agent workflows for such explorative tasks.

In the biomedical domain, \citet{GAO20246125} discussed potential applications of AI agents in brainstorming, experimental planning, and execution. BioResearcher \cite{luo2024intentionimplementationautomatingbiomedical} proposed an end-to-end framework for biomedical research involving dry lab experiments. DrugAgent \cite{liu2025drugagentautomatingaiaideddrug} adopted multi-agent collaboration to automate drug discovery. In chemistry, Coscientist \cite{boiko2023autonomous} incorporated the use of tools by LLMs to support semi-autonomous chemistry experiment design and execution. ProtAgents \cite{ghafarollahi2024protagentsproteindiscoverylarge} facilitated biochemistry discovery by building a multi-agent framework for automating protein design. Recent works, such as FutureHouse \cite{skarlinski2025futurehouse} and AI Co-scientist \cite{gottweis2025aicoscientist}, contributed to formulating demonstrably novel research hypotheses and proposals using multi-agent systems guided by predefined research goals.

\subsection{General Research}
\label{sec:l2-5}

Apart from specialized domain applications, some benchmarks have broadly evaluated diverse tasks from different stages of scientific discovery. DiscoveryWorld \cite{jansen2024discoveryworldvirtualenvironmentdeveloping} created a virtual environment for LLM agents to conduct simplified scientific exploration. In \cite{liu2025visionautoresearchllm}, various application scenarios for AI agents in research were comprehensively discussed, supported by preliminary evaluation datasets. Similarly, CURIE \cite{kon2025curie} proposed a benchmark and an agentic framework for rigorous and automated scientific experimentation. While EAIRA \cite{cappello2025eairaestablishingmethodologyevaluating} focused on assessing the ability of LLMs to perform in a real-world research assistant role using various task formats.
\vspace{-0.1cm}

\section{Level 3. LLM as Scientist (\autoref{tab:level3})}  
\vspace{-0.1cm}

\label{sec:l3}
Recently, several research efforts and commercial products have demonstrated prototypes of fully autonomous research within the artificial intelligence domain. These systems typically encompass a comprehensive workflow, from initial literature review to iterative refinement cycles where hypotheses or designs are progressively improved. A common feature is using an agent-based framework to autonomously produce research outputs, often culminating in draft research papers. This section will further compare these approaches, focusing on their methodologies for idea development and iterative refinement, as these aspects critically distinguish them from \textit{Level 2} agents.
\vspace{-0.1cm}

\subsection{Idea Development}
\vspace{-0.1cm}

The genesis of research in \textit{Level 3} systems involves transforming initial concepts into validated hypotheses, with distinct approaches to sourcing and vetting these ideas. Agent Laboratory \cite{schmidgall2025agentlaboratoryusingllm} predominantly conducts literature reviews based on human-defined research objectives. Moving towards greater autonomy, several systems initiate their process from broader human inputs, such as reference papers \cite{HKUDS2025AIResearcher, Autoscience2025} or general research domains \cite{Intology2025Zochi}, subsequently exploring literature to autonomously identify gaps and formulate novel hypotheses. The AI Scientist (v1 \cite{lu2024aiscientistfullyautomated} and v2 \cite{yamada2025aiscientistv2workshoplevelautomated}) showcases an even more generative approach: v1 brainstorms ideas from templates and past work, while v2 can generate diverse research proposals from abstract thematic prompts. Crucially, these systems employ diverse methods to evaluate their ideas prior to full implementation. AI Scientist-v1 uses self-assessed scores for interestingness, novelty, and feasibility, supplemented by external checks with Semantic Scholar. AI Scientist-v2 integrates literature review tools early in its idea formulation stage to assess novelty. This spectrum reveals a clear trend: while humans often initiate ideas, advanced systems can autonomously explore, generate, and validate the scientific merit and originality of research objectives before development.

\vspace{-0.1cm}

\subsection{Iterative Refinement}
\vspace{-0.1cm}

Iterative refinement within \textit{Level 3} systems involves sophisticated feedback loops that enable not just incremental improvements but also fundamental reassessments of the research trajectory. A key differentiator lies in the primary source and nature of this high-level feedback. The AI Scientist (v1 and v2) incorporates highly automated internal review and refinement processes. It employs AI reviewers, LLM evaluators for experimental choices, and VLMs to critique figures, fostering a rich internal feedback loop for iterative development. In contrast, Zochi \cite{Intology2025Zochi} integrates human expertise for macro-level guidance, where feedback can trigger complete re-evaluations of hypotheses or designs. This allows it to act on critiques challenging the core research premise, even reverting to hypothesis regeneration if results are unsatisfactory. Overall, while automated self-correction is a common goal, the current landscape reveals a pragmatic blend: some systems focus on enhancing autonomous deep reflection, while others integrate human oversight for robust, high-level iterative refinement and strategic redirection.
\vspace{-0.1cm}

\section{Challenges and Future Directions}
\vspace{-0.1cm}

\label{sec:future}
Throughout this survey, we have systematically reviewed the escalating roles of Large Language Models in scientific discovery, delineating their progression through distinct levels of autonomy and capability—from foundational assistants and analysts to increasingly autonomous scientific researchers. In particular, we have underscored the evolving methodologies, task complexities, and the nature of human-LLM interaction that define each stage of this maturation. Beyond reviewing these advancements and current applications, this section presents several significant challenges and outlines promising directions for future research, aiming to inspire further exploration into the development and responsible deployment of LLMs as transformative tools in scientific inquiry.
\vspace{-0.1cm}

\paragraph{Fully-Autonomous Research Cycle}
While current \textit{Level 3} systems can navigate multiple stages of the scientific method for a specific inquiry, they often operate within a single research instance or predefined topic. The scientific method, however, is inherently cyclical, characterized by continuous iteration, refinement, and the pursuit of evolving research questions. A significant future direction, therefore, is to develop LLM-based systems capable of engaging in a truly autonomous research cycle. This would entail not merely executing a given research task from start to finish, but possessing the foresight to discern the broader implications of their findings, proactively identify promising avenues for subsequent investigation, and strategically direct their efforts towards practical advancements that build upon previous work. 
\vspace{-0.1cm}

\paragraph{Robotic Automation}
A key barrier to fully autonomous scientific discovery in natural science is LLM agents' inability to conduct physical laboratory experiments. While adept in computational research, their application in fields requiring physical interaction remains limited. Integrating LLMs with robotic systems empowers them to translate their planning capabilities into direct experimental actions. Early works in LLM-robotic integration \cite{yoshikawa2023large, song2024multiagent, darvish2025organaroboticassistantautomated} already highlights this potential in chemical experimentation. Such automation is poised to significantly broaden LLM-based research, enabling end-to-end discovery in disciplines like chemistry and materials science, thereby advancing autonomous scientific exploration.

\vspace{-0.1cm}

\paragraph{Transparency and Interpretability}
The \textit{black-box} nature (or opacity) of advancing LLMs in science undermines scientific validation, trust, and the assimilation of AI-driven insights \cite{xu2025advancingaiscientistunderstandingmaking}. Addressing this opacity demands more than superficial Explainable AI (XAI) techniques \cite{10.1007/978-3-031-67751-9_1}. It necessitates a paradigm shift towards systems whose internal operations are inherently designed for verifiable reasoning and justifiable conclusions \cite{bengio2025superintelligentagentsposecatastrophic}. Consequently, the challenge is not just explaining outputs, but ensuring the AI's internal logic aligns with scientific principles and can reliably differentiate asserted claims from verifiable truths. This profound interpretability is vital for reliable and reproducible LLM-based scientific discovery.

\vspace{-0.1cm}

\paragraph{Continuous Self-Improvement}

The iterative and evolving nature of scientific inquiry demands systems capable of learning from ongoing engagement, assimilating experimental outcomes, and adapting research strategies. Current research integrating continual learning with agent-based systems already highlights the potential for LLMs to adapt to new tasks or environments without catastrophic forgetting \cite{majumder2023clincontinuallylearninglanguage, kim2024onlinecontinuallearninginteractive}. Within scientific discovery, a promising future direction is to incorporate online reinforcement learning frameworks \cite{pmlr-v202-carta23a}. This integration promises to continuously enhance scientific agents' capabilities over their operational lifetime through successive discoveries, thereby advancing sustainable autonomous exploration.
\vspace{-0.1cm}

\paragraph{Ethics and Societal Alignment}

As LLM-based systems gain independent reasoning and action capabilities, their potential for risks—ranging from amplified societal biases to deliberate misuse like generating harmful substances or challenging human control—becomes increasingly salient and complex \cite{he2023controlriskpotentialmisuse, 10.1007/978-3-031-67751-9_1, bengio2025superintelligentagentsposecatastrophic}. With AI capabilities and societal norms in constant flux, alignment is consequently an imperative, continuous process demanding adaptive governance and evolving value systems \cite{li2024agentalignmentevolvingsocial}. This requires embedding ethical constraints directly in scientific AI design frameworks, alongside vigilant oversight, to ensure advancements serve human well-being and the common good.

\newpage
\section*{Limitations}

This survey provides a systematic review of LLMs in scientific discovery, with a particular emphasis on the paradigm shift characterized by their escalating levels of autonomy. Our analysis and the selection of reviewed literature are therefore centered on works that illustrate this transition across the stages of the scientific method, categorized within our proposed three-level autonomy framework: LLM as Tool, LLM as Analyst, and LLM as Scientist.

Consequently, the scope of this survey has certain limitations. Firstly, we do not provide an exhaustive review of research focused on the development of general-purpose scientific LLMs for domain-specific reasoning or application. These areas, while crucial to the broader landscape of AI in science, are extensively covered in other existing surveys and fall outside our specific focus on the autonomy paradigm. Secondly, while we acknowledge the importance of fundamental LLM capabilities such as planning, code generation, and agentic decision-making, this survey does not delve deeply into orthogonal benchmarks or methodologies related to these general abilities. These exclusions were deliberate to maintain a focused exploration of the transformative roles and increasing independence of LLMs throughout the scientific research lifecycle.

\section*{Ethics Statement}
Our paper presents a comprehensive survey of LLMs in scientific discovery, with a specific focus on their role transformation from task automation tools to autonomous agents. All research works reviewed in this survey are properly cited. To the best of our knowledge, the referenced materials are publicly accessible or available under licenses permitting their use for research review. We did not conduct additional dataset curation or human annotation work. Consequently, we believe that this paper does not raise any ethical concerns.

\section*{Acknowledgements}

We thank all the anonymous reviewers and meta reviewers for their valuable comments. The authors of this paper were supported by the ITSP Platform Research Project (ITS/189/23FP) from ITC of Hong Kong, SAR, China, and the AoE (AoE/E-601/24-N), the RIF (R6021-20) and the GRF (16205322) from RGC of Hong Kong, SAR, China.

\bibliography{main}

\setcounter{table}{0}
\renewcommand{\thetable}{A\arabic{table}}

\setcounter{figure}{0}
\renewcommand{\thefigure}{A\arabic{figure}}

\newpage
\onecolumn

\appendix

\section{Summary Tables of LLMs in Scientific Discovery}
\label{sec:tables}

\begin{table*}[!h]
\centering
\scriptsize
\begin{tabular}{lccccc}
\toprule
\textbf{Research Works} & \textbf{Science Domain} & \textbf{Task Nature} & \textbf{\begin{tabular}[c]{@{}c@{}}Framework\\Methodology\end{tabular}} & \textbf{\begin{tabular}[c]{@{}c@{}}Evaluation\\Benchmark\end{tabular}} & \textbf{\begin{tabular}[c]{@{}c@{}}Agentic\\Workflow\end{tabular}} \\ \midrule
\multicolumn{6}{c}{\textit{Literature Search and Info Aggregation}} \\
\rowcolor[HTML]{DAE8FC}
LitLLM{\tiny ~\cite{agarwal2024litllmtoolkitscientificliterature}} & General & Literature & \ding{51} & \ding{55} & \ding{55} \\
Science Hierarchography{\tiny ~\cite{gao2025sciencehierarchographyhierarchicalorganization}}  & General & Literature & \ding{51} & \ding{55} & \ding{51} \\
\rowcolor[HTML]{DAE8FC}\citet{Dennstadt2024Title} & Biomedicine & Literature & \ding{51} & \ding{55} & \ding{55} \\

SCIMON{\tiny ~\cite{wang2024scimonscientificinspirationmachines}} & General & Literature, Idea Generation & \ding{51} & \ding{55} & \ding{55} \\
\rowcolor[HTML]{DAE8FC}ResearchAgent{\tiny ~\cite{baek2025researchagentiterativeresearchidea}} & General & Literature, Idea Generation & \ding{51} & \ding{55} & \ding{51} \\

Text-Tuple-Table{\tiny ~\cite{deng2024texttupletableinformationintegrationtexttotable}} & General & Text2Table & \ding{51} & \ding{51} & \ding{55} \\
\rowcolor[HTML]{DAE8FC}TKGT{\tiny ~\cite{jiang-etal-2024-tkgt}} & General & Text2Table & \ding{51} & \ding{51} & \ding{55} \\

ArxivDIGESTables{\tiny ~\cite{newman2024arxivdigestablessynthesizingscientificliterature}} & General & Literature, Text2Table & \ding{51} & \ding{51} & \ding{55} \\
\rowcolor[HTML]{DAE8FC}arXiv2Table{\tiny ~\cite{wang2025llmsgeneratetabularsummaries}} & General & Literature, Text2Table & \ding{51} & \ding{51} & \ding{55} \\ 

PaperQA \& LitQA{\tiny ~\cite{Lala2023PaperQA}}  & General & Literature & \ding{51} & \ding{51} & \ding{55} \\
\rowcolor[HTML]{DAE8FC}AutoSurvey{\tiny ~\cite{wang2024autosurvey}} & General & Literature & \ding{51} & \ding{55} & \ding{51} \\

\midrule

\multicolumn{6}{c}{\textit{Idea Generation and Hypothesis Formulation}} \\
\rowcolor[HTML]{DAE8FC}
\citet{si2024llmsgeneratenovelresearch} & Artificial Intelligence & Idea Generation & \ding{55} & \ding{51} & \ding{55} \\
LiveIdeaBench{\tiny ~\cite{ruan2025liveideabenchevaluatingllmsdivergent}} & General & Idea Generation & \ding{55} & \ding{51} & \ding{55} \\
\rowcolor[HTML]{DAE8FC}
Nova{\tiny ~\cite{hu2024novaiterativeplanningsearch}} & General & Literature, Idea Generation & \ding{51} & \ding{55} & \ding{51} \\
IdeaBench{\tiny ~\cite{guo2024ideabenchbenchmarkinglargelanguage}} & General & Literature, Idea Generation & \ding{55} & \ding{51} & \ding{55} \\
\rowcolor[HTML]{DAE8FC}
GraphEval{\tiny ~\cite{feng2025graphevallightweightgraphbasedllm}} & Artificial Intelligence & Literature, Idea Generation & \ding{55} & \ding{51} & \ding{55} \\
AI Idea Bench 2025{\tiny ~\cite{qiu2025aiideabench2025}} & Artificial Intelligence & Literature, Idea Generation & \ding{55} & \ding{51} & \ding{55} \\
\rowcolor[HTML]{DAE8FC}
\citet{buehler2024acceleratingscientificdiscoverygenerative} & Biology & Literature, Idea Generation & \ding{51} & \ding{55} & \ding{55} \\
SciAgents{\tiny ~\cite{ghafarollahi2024sciagentsautomatingscientificdiscovery}} & General & Literature, Idea / Hypothesis Generation & \ding{51} & \ding{55} & \ding{51} \\ 
\rowcolor[HTML]{DAE8FC}
MOOSE-Chem{\tiny ~\cite{yang2025moosechemlargelanguagemodels}} & Chemistry & Literature, Idea / Hypothesis Generation & \ding{51} & \ding{51} & \ding{51} \\
\citet{yang2024largelanguagemodelsautomated} & General & Literature, Idea / Hypothesis Generation & \ding{55} & \ding{51} & \ding{55} \\
\rowcolor[HTML]{DAE8FC}
ResearchBench{\tiny ~\cite{liu2025researchbenchbenchmarkingllmsscientific}} & General & Literature, Idea / Hypothesis Generation & \ding{55} & \ding{51} & \ding{55} \\
KG-CoI{\tiny ~\cite{xiong2024improvingscientifichypothesisgeneration}}& General & Literature, Idea / Hypothesis Generation & \ding{51} & \ding{55} & \ding{51}\\
\rowcolor[HTML]{DAE8FC}
HypER{\tiny ~\cite{vasu2025hyper}} & General & Literature, Idea / Hypothesis Generation & \ding{51} & \ding{55} & \ding{51} \\
\citet{oneill2025sparks} & General & Literature, Idea / Hypothesis Generation & \ding{51} & \ding{55} & \ding{51} \\
\rowcolor[HTML]{DAE8FC}\citet{ciucă2023harnessingpoweradversarialprompting} & Astronomy & Hypothesis Generation & \ding{51} & \ding{55} & \ding{55} \\
\citet{D3DD00185G} & Biomedicine & Hypothesis Generation & \ding{51} & \ding{55} & \ding{55} \\
\rowcolor[HTML]{DAE8FC}LLM4GRN{\tiny ~\cite{afonja2024llm4grndiscoveringcausalgene}} & Biology & Hypothesis Generation & \ding{51} & \ding{55} & \ding{55} \\
\citet{Zhou_2024} & General & Hypothesis Generation & \ding{51} & \ding{51} & \ding{55} \\
\rowcolor[HTML]{DAE8FC}\citet{qi2023largelanguagemodelszero} & General & Hypothesis Generation & \ding{55} & \ding{51} & \ding{55} \\
\citet{qi2024largelanguagemodelsbiomedical} & Biomedicine & Hypothesis Generation & \ding{55} & \ding{51} & \ding{55} \\
\rowcolor[HTML]{DAE8FC}Scideator{\tiny ~\cite{radensky2025scideatorhumanllmscientificidea}} & General & Idea / Hypothesis Generation & \ding{51} & \ding{55} & \ding{55}\\
\citet{li2024learning} & General & Idea / Hypothesis Generation & \ding{51} & \ding{55} & \ding{55} 
\\

\midrule

\multicolumn{6}{c}{\textit{Experiment Planning and Execution}} \\
\rowcolor[HTML]{DAE8FC}\citet{li2025largelanguagemodelshelp} & General & Planning & \ding{51} & \ding{55} & \ding{55} \\
\citet{shi2025hierarchicallyencapsulatedrepresentationprotocol} & Biology & Planning & \ding{51} & \ding{55} & \ding{51} \\
\rowcolor[HTML]{DAE8FC}BioPlanner{\tiny ~\cite{odonoghue2023bioplannerautomaticevaluationllms}} & Biology & Planning & \ding{55} & \ding{51} & \ding{55} \\
ARCADE{\tiny ~\cite{yin2022naturallanguagecodegeneration}} & Artificial Intelligence & Code Generation & \ding{51} & \ding{51} & \ding{55} \\
\rowcolor[HTML]{DAE8FC}AIDE{\tiny ~\cite{jiang2025aideaidrivenexplorationspace}} & Artificial Intelligence & Code Generation & \ding{51} & \ding{55} & \ding{51} \\
SciCode{\tiny ~\cite{tian2024scicoderesearchcodingbenchmark}} & Artificial Intelligence & Code Generation & \ding{55} & \ding{51} & \ding{55} \\
\rowcolor[HTML]{DAE8FC}DS-1000{\tiny ~\cite{lai2022ds1000naturalreliablebenchmark}} & Artificial Intelligence & Code Generation & \ding{55} & \ding{51} & \ding{55} \\
MLE-Bench{\tiny ~\cite{chan2025mlebenchevaluatingmachinelearning}} & Artificial Intelligence & Code Generation & \ding{55} & \ding{51} & \ding{51} \\  \midrule
\multicolumn{6}{c}{\textit{Data Analysis and Organization}} \\

\rowcolor[HTML]{DAE8FC}AutomaTikZ{\tiny ~\cite{belouadi2024automatikztextguidedsynthesisscientific}} & General & Text2Chart & \ding{51} & \ding{51} & \ding{55} \\
Text2Chart31{\tiny ~\cite{zadeh2025text2chart31instructiontuningchart}} & General & Text2Chart & \ding{51} & \ding{55} & \ding{55} \\
\rowcolor[HTML]{DAE8FC}ChartX \& ChartVLM{\tiny ~\cite{xia2025chartxchartvlmversatile}} & General & Chart Reasoning & \ding{51} & \ding{51} & \ding{55} \\
CharXiv{\tiny ~\cite{wang2024charxivchartinggapsrealistic}} & General & Chart Reasoning & \ding{55} & \ding{51} & \ding{55} \\
\rowcolor[HTML]{DAE8FC}ChartQA{\tiny ~\cite{masry2022chartqabenchmarkquestionanswering}} & General & Chart Reasoning & \ding{55} & \ding{51} & \ding{55} \\
Chain-of-Table{\tiny ~\cite{wang2024chainoftableevolvingtablesreasoning}} & General & Tabular Reasoning & \ding{51} & \ding{55} & \ding{51} \\
\rowcolor[HTML]{DAE8FC}TableBench{\tiny ~\cite{wu2025tablebenchcomprehensivecomplexbenchmark}} & General & Tabular Reasoning & \ding{55} & \ding{51} & \ding{55} \\
\citet{deng2024tablestextsimagesevaluating} & General & Tabular Reasoning & \ding{55} & \ding{51} & \ding{55} \\ \midrule

\multicolumn{6}{c}{\textit{Conclusion and Hypothesis Validation}} \\

\rowcolor[HTML]{DAE8FC}
\citet{tyser2024aidrivenreviewsystemsevaluating} & General & Review & \ding{51} & \ding{55} & \ding{55} \\
ClaimCheck{\tiny ~\cite{ou2025claimcheckgroundedllmcritiques}} & Artificial Intelligence & Review & \ding{55} & \ding{51} & \ding{55} \\
\rowcolor[HTML]{DAE8FC}
\citet{du2024llmsassistnlpresearchers} & Artificial Intelligence & Review & \ding{55} & \ding{51} & \ding{55} \\
\citet{zhou-etal-2024-llm} & Artificial Intelligence & Review & \ding{55} & \ding{51} & \ding{55} \\
\rowcolor[HTML]{DAE8FC}
ReviewerGPT{\tiny ~\cite{liu2023reviewergptexploratorystudyusing}} & Artificial Intelligence & Review & \ding{55} & \ding{51} & \ding{55} \\
\citet{bougie2024generative} & General & Review & \ding{51} & \ding{55} & \ding{51} \\
\rowcolor[HTML]{DAE8FC}
CycleResearcher{\tiny ~\cite{weng2025cycleresearcherimprovingautomatedresearch}} & Artificial Intelligence & Review & \ding{51} & \ding{51} & \ding{51} \\
\citet{takagi2023autonomoushypothesisverificationlanguage} & General & Hypothesis Validation & \ding{51} & \ding{55} & \ding{55} \\
\rowcolor[HTML]{DAE8FC}
\citet{wen2025predicting} & Artificial Intelligence & Hypothesis Validation & \ding{51} & \ding{55} & \ding{55} \\
PaperBench{\tiny ~\cite{starace2025paperbench}} & Artificial Intelligence & Hypothesis Validation & \ding{55} & \ding{51} & \ding{51} \\ 
\rowcolor[HTML]{DAE8FC}
SciReplicate-Bench{\tiny ~\cite{xiang2025scireplicatebenchbenchmarkingllmsagentdriven}} & Artificial Intelligence & Hypothesis Validation & \ding{55} & \ding{51} & \ding{51} \\
\citet{xu2025advancingaiscientistunderstandingmaking} & Physics & Hypothesis Validation & \ding{51} & \ding{55} & \ding{51} \\\midrule

\multicolumn{6}{c}{\textit{Iteration and Refinement}} \\

\rowcolor[HTML]{DAE8FC}\citet{quan2024verificationrefinementnaturallanguage} & General & Refinement & \ding{51} & \ding{55} & \ding{55} \\
MC-NEST{\tiny ~\cite{rabby2025iterativehypothesisgenerationscientific}} & General & Hypothesis Generation, Refinement & \ding{51} & \ding{55} & \ding{51} \\
\rowcolor[HTML]{DAE8FC}Chain of Ideas{\tiny ~\cite{li2024chainideasrevolutionizingresearch}} & Artificial Intelligence & Idea Generation, Refinement & \ding{51} & \ding{51} & \ding{51} \\

\bottomrule
\end{tabular}
\caption{Comparison and classification of \textit{Level 1} research works in LLM-based scientific discovery.}
\label{tab:level1}
\end{table*}

\begin{table*}[!h]
\centering
\scriptsize
\begin{tabular}{lccccccccc}
\toprule
\multirow{2}{*}{\textbf{Research Works}} & \multirow{2}{*}{\textbf{Science Domain}} & \multirow{2}{*}{\textbf{\begin{tabular}[c]{@{}c@{}}Methodology\\ Framework\end{tabular}}} & \multirow{2}{*}{\textbf{\begin{tabular}[c]{@{}c@{}}Benchmark\\ Evaluation\end{tabular}}} & \multicolumn{6}{c}{\textbf{Scientific Method Stages}} \\ \cmidrule{5-10} 
 &  &  &  & \textit{Obs.} & \textit{Hyp.} & \textit{Exp.} & \textit{Ana.} & \textit{Con.} & \textit{Ref.} \\ \midrule

\multicolumn{10}{c}{\textit{Machine Learning Research}} \\

\rowcolor[HTML]{DAFCD4}CodeScientist{\tiny ~\cite{jansen2025codescientistendtoendsemiautomatedscientific}} & Artificial Intelligence & \ding{51} & \ding{55} & \ding{51} & \ding{51} & \ding{51} & \ding{55} & \ding{55} & \ding{55} \\
BudgetMLAgent{\tiny ~\cite{gandhi2025budgetmlagentcosteffectivellmmultiagent}} & Artificial Intelligence & \ding{51} & \ding{55} & \ding{55} & \ding{51} & \ding{51} & \ding{51} & \ding{51} & \ding{55} \\
\rowcolor[HTML]{DAFCD4}IMPROVE{\tiny ~\cite{xue2025improveiterativemodelpipeline}} & Artificial Intelligence & \ding{51} & \ding{55} & \ding{55} & \ding{55} & \ding{51} & \ding{51} & \ding{51} & \ding{51} \\
MLAgentBench{\tiny ~\cite{huang2024mlagentbenchevaluatinglanguageagents}} & Artificial Intelligence & \ding{55} & \ding{51} & \ding{55} & \ding{55} & \ding{51} & \ding{51} & \ding{51} & \ding{55} \\ 
\rowcolor[HTML]{DAFCD4}MLR-Copilot{\tiny ~\cite{li2024mlrcopilot}} & Artificial Intelligence & \ding{51} & \ding{55} & \ding{55} & \ding{55} & \ding{51} & \ding{51} & \ding{55} & \ding{51} \\
MLRC-Bench{\tiny ~\cite{zhang2025mlrcbench}} & Artificial Intelligence & \ding{55} & \ding{51} & \ding{55} & \ding{55} & \ding{51} & \ding{51} & \ding{55} & \ding{51} \\
\rowcolor[HTML]{DAFCD4}RE-Bench{\tiny ~\cite{wijk2024rebench}} & Artificial Intelligence & \ding{55} & \ding{51} & \ding{55} & \ding{55} & \ding{51} & \ding{51} & \ding{55} & \ding{51} \\
MLZero{\tiny ~\cite{fang2025mlzero}} & Artificial Intelligence & \ding{51} & \ding{51} & \ding{55} & \ding{55} & \ding{51} & \ding{51} & \ding{55} & \ding{51} \\
\rowcolor[HTML]{DAFCD4}Genesys{\tiny ~\cite{cheng2025language}} & Artificial Intelligence & \ding{51} & \ding{55} & \ding{55} & \ding{55} & \ding{51} & \ding{55} & \ding{55} & \ding{51} \\
MLGym{\tiny ~\cite{nathani2025mlgym}} & Artificial Intelligence & \ding{51} & \ding{51} & \ding{55} & \ding{55} & \ding{51} & \ding{51} & \ding{55} & \ding{51} \\ \midrule

\multicolumn{10}{c}{\textit{Data Modeling and Analysis}} \\

\rowcolor[HTML]{DAFCD4}DAgent{\tiny ~\cite{xu2025dagentrelationaldatabasedrivendata}} & Data Science & \ding{51} & \ding{55} & \ding{55} & \ding{51} & \ding{51} & \ding{51} & \ding{51} & \ding{55} \\
DS-Agent{\tiny ~\cite{guo2024dsagentautomateddatascience}} & Data Science & \ding{51} & \ding{55} & \ding{55} & \ding{51} & \ding{51} & \ding{51} & \ding{51} & \ding{51} \\

\rowcolor[HTML]{DAFCD4}InfiAgent-DABench{\tiny ~\cite{hu2024infiagentdabenchevaluatingagentsdata}} & Data Science & \ding{55} & \ding{51} & \ding{55} & \ding{51} & \ding{51} & \ding{51} & \ding{55} & \ding{55} \\
BLADE{\tiny ~\cite{gu2024bladebenchmarkinglanguagemodel}} & Data Science & \ding{55} & \ding{51} & \ding{55} & \ding{55} & \ding{51} & \ding{51} & \ding{51} & \ding{55} \\
\rowcolor[HTML]{DAFCD4}DiscoveryBench{\tiny ~\cite{majumder2024discoverybenchdatadrivendiscoverylarge}} & Data Science & \ding{55} & \ding{51} & \ding{55} & \ding{51} & \ding{51} & \ding{51} & \ding{51} & \ding{55} \\
DSBench{\tiny ~\cite{jing2024dsbench}} & Data Science & \ding{55} & \ding{51} & \ding{55} & \ding{55} & \ding{51} & \ding{51} & \ding{51} & \ding{55} \\

\rowcolor[HTML]{DAFCD4}\citet{zheng2023largelanguagemodelsscientific} & General & \ding{51} & \ding{51} & \ding{51} & \ding{51} & \ding{51} & \ding{51} & \ding{51} & \ding{55} \\
\midrule

\multicolumn{10}{c}{\textit{Function Discovery}} \\
\rowcolor[HTML]{DAFCD4}BoxLM{\tiny ~\cite{li2024automatedstatisticalmodeldiscovery}} & Statistics & \ding{51} & \ding{55} & \ding{55} & \ding{51} & \ding{51} & \ding{51} & \ding{51} & \ding{55} \\
LLM-SR{\tiny ~\cite{shojaee2025llmsrscientificequationdiscovery}} & General & \ding{51} & \ding{55} & \ding{55} & \ding{51} & \ding{51} & \ding{55} & \ding{51} & \ding{51} \\
\rowcolor[HTML]{DAFCD4}LLM-SRBench{\tiny ~\cite{shojaee2025llmsrbenchnewbenchmarkscientific}} & General & \ding{55} & \ding{51} & \ding{55} & \ding{51} & \ding{51} & \ding{55} & \ding{51} & \ding{51} \\
Gravity-Bench-v1{\tiny ~\cite{koblischke2025gravitybenchv1benchmarkgravitationalphysics}} & Physics & \ding{55} & \ding{51} & \ding{55} & \ding{51} & \ding{51} & \ding{55} & \ding{51} & \ding{51} \\
\rowcolor[HTML]{DAFCD4}DrSR{\tiny ~\cite{wang2025drsr}} & General & \ding{51} & \ding{55} & \ding{55} & \ding{51} & \ding{51} & \ding{51} & \ding{51} & \ding{51} \\
EvoSLD{\tiny ~\cite{wei2025evosld}} & Artificial Intelligence & \ding{51} & \ding{55} & \ding{55} & \ding{51} & \ding{51} & \ding{51} & \ding{51} & \ding{51} \\ \midrule

\multicolumn{10}{c}{\textit{Natural Science Research}} \\

\rowcolor[HTML]{DAFCD4}Coscientist{\tiny ~\cite{boiko2023autonomous}}  & Chemistry & \ding{51} & \ding{55} & \ding{55} & \ding{51} & \ding{51} & \ding{51} & \ding{51} & \ding{55} \\
\citet{GAO20246125}& Biomedicine & \ding{51} & \ding{55} & \ding{55} & \ding{51} & \ding{51} & \ding{51} & \ding{51} & \ding{55} \\
\rowcolor[HTML]{DAFCD4}BioResearcher{\tiny ~\cite{luo2024intentionimplementationautomatingbiomedical}} & Biomedicine & \ding{51} & \ding{55} & \ding{55} & \ding{55} & \ding{51} & \ding{51} & \ding{51} & \ding{51} \\
DrugAgent{\tiny ~\cite{liu2025drugagentautomatingaiaideddrug}} & Biomedicine & \ding{51} & \ding{55} & \ding{55} & \ding{51} & \ding{51} & \ding{55} & \ding{55} & \ding{51} \\

\rowcolor[HTML]{DAFCD4}FutureHouse{\tiny ~\cite{skarlinski2025futurehouse}} & Chemistry, Biology &  \ding{51} & \ding{55} & \ding{51} & \ding{51} & \ding{51} & \ding{55} & \ding{55} & \ding{55} \\
ScienceAgentBench{\tiny ~\cite{chen2025scienceagentbenchrigorousassessmentlanguage}}  & Chemistry, Biology & \ding{55} & \ding{51} & \ding{51} & \ding{51} & \ding{51} & \ding{51} & \ding{51} & \ding{55} \\

\rowcolor[HTML]{DAFCD4}ProtAgents{\tiny ~\cite{ghafarollahi2024protagentsproteindiscoverylarge}} & Chemistry, Biology & \ding{51} & \ding{55} & \ding{51} & \ding{51} & \ding{51} & \ding{51} & \ding{51} & \ding{55} \\

Auto-Bench{\tiny ~\cite{chen2025autobenchautomatedbenchmarkscientific}}  & General & \ding{55} & \ding{51} & \ding{55} & \ding{55} & \ding{51} & \ding{51} & \ding{55} & \ding{51}\\
\rowcolor[HTML]{DAFCD4}AI co-scientist{\tiny ~\cite{gottweis2025aicoscientist}} & General & \ding{51} & \ding{55} & \ding{55} & \ding{51} & \ding{51} & \ding{51} & \ding{51} & \ding{55} \\

\midrule

\multicolumn{10}{c}{\textit{General Research}} \\

\rowcolor[HTML]{DAFCD4}DiscoveryWorld{\tiny ~\cite{jansen2024discoveryworldvirtualenvironmentdeveloping}} & General & \ding{55} & \ding{51} & \ding{55} & \ding{51} & \ding{51} & \ding{51} & \ding{51} & \ding{51} \\
\citet{liu2025visionautoresearchllm} & General & \ding{55} & \ding{51} & \ding{51} & \ding{51} & \ding{51} & \ding{51} & \ding{51} & \ding{51} \\
\rowcolor[HTML]{DAFCD4}Curie{\tiny ~\cite{kon2025curie}} & General &  \ding{51} & \ding{55} & \ding{55} & \ding{55} & \ding{51} & \ding{51} & \ding{51} & \ding{55} \\
EAIRA{\tiny ~\cite{cappello2025eairaestablishingmethodologyevaluating}}  & General & \ding{55} & \ding{51} & \ding{51} & \ding{51} & \ding{51} & \ding{51} & \ding{51} & \ding{55} \\

\bottomrule

\end{tabular}
\caption{Comparison and classification of \textit{Level 2} research works in LLM-based scientific discovery.}
\label{tab:level2}
\end{table*}

\setlength{\tabcolsep}{2.5pt}
\begin{table*}[!h]
\centering
\scriptsize
\begin{tabular}{lccccc}
\toprule
\multirow{2}{*}{\textbf{Research Works}} & \multicolumn{1}{c}{\multirow{2}{*}{\textbf{Science Domain}}} & \multirow{2}{*}{\textbf{\begin{tabular}[c]{@{}c@{}}Methodology\\ Framework\end{tabular}}} & \multirow{2}{*}{\textbf{\begin{tabular}[c]{@{}c@{}}Benchmark\\ Evaluation\end{tabular}}} & \multirow{2}{*}{\textbf{\begin{tabular}[c]{@{}c@{}}Featured\\ Functionality\end{tabular}}} & \multirow{2}{*}{\textbf{Open-Sourced?}} \\
 & \multicolumn{1}{c}{} &  &  &  &  \\ \midrule
 
\rowcolor[HTML]{F5F4C9}Agent Laboratory{\tiny ~\cite{schmidgall2025agentlaboratoryusingllm}} & Artificial Intelligence & \ding{51} & \ding{51} & \begin{tabular}[c]{@{}c@{}}literature review, experimentation, report writing,\\ iterative research with human feedback loops.\end{tabular} & \ding{51} \\

The AI Scientist{\tiny ~\cite{lu2024aiscientistfullyautomated}} & Artificial Intelligence & \ding{51} & \ding{55} & \begin{tabular}[c]{@{}c@{}}idea generation, code generation, \\ experiment execution, research paper writing.\end{tabular} & \ding{51} \\

\rowcolor[HTML]{F5F4C9}The AI Scientist-v2{\tiny ~\cite{yamada2025aiscientistv2workshoplevelautomated}} & Artificial Intelligence & \ding{51} & \ding{55} & \begin{tabular}[c]{@{}c@{}}idea generation, code generation, \\ experiment execution, research paper writing,\\ with agentic tree-search and feedbacks.\end{tabular} & \ding{51} \\

AI-Researcher{\tiny ~\cite{HKUDS2025AIResearcher}} & Artificial Intelligence & \ding{51} & \ding{55} & \begin{tabular}[c]{@{}c@{}}literature review, data analysis,\\ report generation.\end{tabular} & \ding{51} \\

\rowcolor[HTML]{F5F4C9}Zochi{\tiny ~\cite{Intology2025Zochi}} & Artificial Intelligence & \ding{51} & \ding{55} & \begin{tabular}[c]{@{}c@{}}customizable workflows for data collection,\\ analysis, and decision-making.\end{tabular} & \ding{51} \\

Carl{\tiny ~\cite{Autoscience2025}} & Artificial Intelligence & \ding{51} & \ding{55} & \begin{tabular}[c]{@{}c@{}}hypothesis generation, experiment design,\\ data analysis, and manuscript writing.\end{tabular} & \ding{55} \\

 \bottomrule
\end{tabular}

\caption{Comparison and classification of \textit{Level 3} research works in LLM-based scientific discovery.}
\label{tab:level3}
\end{table*}

\end{document}